\documentclass[sigconf]{acmart}
\AtBeginDocument{%
  \providecommand\BibTeX{{%
    \normalfont B\kern-0.5em{\scshape i\kern-0.25em b}\kern-0.8em\TeX}}}

\setcopyright{acmcopyright}
\copyrightyear{2022}
\acmYear{2022}
\acmDOI{}

\acmConference[SIGIR '22 PatentSemTech]{SIGIR 3rd Workshop on Patent Text Mining and Semantic Technologies}{July 15,
  2022}{Madrid, Spain}
\acmPrice{}
\acmISBN{}

\begin{document}

\title{Recent Developments in AI and USPTO Open Data}

\author{Scott Beliveau}
\authornote{The views expressed in this extended abstract should not be construed as official policy statements of the United States Patent and Trademark Office or of the U.S. Government. All errors are the authors' own.}
\email{Scott.Beliveau@uspto.gov}
\affiliation{%
  \institution{United States Patent and Trademark Office}
  \city{Alexandria}
  \state{Virginia}
  \country{USA}
}

\author{Jerry Ma}
\authornotemark[1]
\email{Jerry.Ma@uspto.gov}
\orcid{0000-0003-4853-0724}
\affiliation{%
  \institution{United States Patent and Trademark Office}
  \city{Alexandria}
  \state{Virginia}
  \country{USA}
}

\begin{abstract}

The USPTO disseminates one of the largest publicly accessible repositories of scientific, technical, and commercial data worldwide. USPTO data has historically seen frequent use in fields such as patent analytics, economics, and prosecution \& litigation tools. This article highlights an emerging class of usecases directed to the research, development, and application of artificial intelligence technology. Such usecases contemplate both the delivery of artificial intelligence capabilities for practical IP applications and the enablement of future state-of-the-art artificial intelligence research via USPTO data products. Examples from both within and beyond the USPTO are offered as case studies.

\end{abstract}

\maketitle

\section{Introduction}

As America's national IP office, the United States Patent and Trademark Office (USPTO) is charged with the mission of granting patents and registering trademarks. The USPTO is required by law to disseminate most nonprovisional patent applications and granted patents to the public.~\footnote{35 U.S.C. § 122(b); 37 C.F.R. § 1.11(a).} U.S. patent data has long been used in many domains of application, including in patent analytics~\citep{oldham2016}, economics~\citep{bell2018, akcigit2018}, and commercial tools for patent prosecutors and litigators,~\footnote{Because agency practice is to refrain from public endorsements of any particular commercial IP products, we omit references to specific examples.} thus serving as a versatile substrate for illustrating the dynamics of national and global innovation.

Concurrently, the fields of artificial intelligence (AI) and natural language processing (NLP) have witnessed a remarkable cadence of scientific breakthroughs. Fueled by model architecture innovations such as self-attention~\citep{vaswani2017} and by the ever-increasing computational horsepower of the leading AI hardware accelerators~\citep{svedin2021,wang2019}, these novel techniques have found versatile areas of application, from machine translation~\citep{vaswani2017} to structural biology~\citep{rives2021}.

In this article, we survey recent developments at the intersection of AI and USPTO data. These developments fall in two broad categories:
\begin{enumerate}
    \item Promising AI and NLP techniques can be brought to bear on USPTO data in existing or novel fields of application.
    \item USPTO data can contribute to work that advances the frontiers of AI and NLP research.
\end{enumerate}

Both bodies of work hold great promise for advancing scientific and technical progress. We encourage those in both the AI and the IP communities to explore how USPTO data can unlock numerous exciting opportunities---both in their respective disciplines and at the intersection of AI \& IP.
\section{USPTO data for patent-focused AI \& NLP usecases}

One distinct body of work uses AI \& NLP techniques on USPTO patent data toward enhancing the value of such data toward longstanding areas of application. Here, we present case studies in the context of IP administration, practice, and empirical research.

\subsection{AI \& NLP tools for IP administration} \label{sec:ai-nlp-ip-administration}

IP offices worldwide seek to apply AI \& NLP in the administration of their respective IP systems, and the USPTO is no exception. The USPTO recognizes the advent of AI as among the most consequential technologies---both for global society as a whole and for the agency's mission of delivering reliable, timely, and quality IP rights~\citep{hirshfeld2021}.

Operationally, the USPTO focuses on two critical areas of AI application: prior art search and patent classification. AI is a natural tool with which to augment prior art search systems. Representation learning and related techniques can produce semantically meaningful embeddings of language, graphs, images, and even proteins~\citep{devlin2019, dwivedi2021, dosovitskiy2021, rives2021}. The USPTO applies such techniques on the agency's patent archives and uses the results toward improving examiner-facing search systems to surface more relevant prior art documents~\citep{uspto2022}.

Turning to patent classification, the USPTO currently classifies patents using a two-stage process. First, the agency assigns a set of Cooperative Patent Classification (CPC) symbols to each patent to characterize the relevant technologies contained therein. Second, the agency determines the subset of CPC symbols (``claim indicators'') associated with claim scope. The USPTO has recently deployed an AI system, trained on annotated USPTO patent data, for assigning claim indicators, and the agency is currently augmenting the system to assign the full set of CPC symbols~\citep{hirshfeld2021}.

\subsection{AI \& NLP tools for IP practitioners \& inventors}

Since the dawn of computer-based information retrieval, software developers have built tools for assisting IP practitioners and inventors. Some tools are similar to those needed by IP offices (\emph{e.g.}, prior art search), while others are specific to the needs of the private IP bar and inventors (\emph{e.g.}, IP portfolio intelligence). Recent work has used publicly-available USPTO data to train AI models that provide new or enhanced capabilities to IP software products.

The USPTO has recently released AI-empowered search capabilities to the public through the Inventor Search Assistant~\citep{inventorsdigest2022}. This tool surfaces not only published applications from the USPTO patent archives, but also non-patent literature (NPL) and foreign patent documents. Traditional prior art search systems have a steep learning curve (\emph{e.g.}, basic query syntax, proximity operators) that may pose a hurdle to early-stage and independent inventors. Such inventors can especially benefit from the Inventor Search Assistant, which uses machine learning techniques to offer an initial overview of the state-of-the-art from natural language queries alone.

\subsection{AI-powered empirical research \& analytics}

Finally, USPTO data can be elucidated via existing AI \& NLP techniques to produce boundary-pushing empirical research \& analytics. A common patent analysis task is to sort patent documents into specific fields of technology or business applications---commonly known as ``patent landscaping''~\citep{trippe2015}. Recent work has applied deep learning to the task of patent landscaping~\citep{abood2018}, with USPTO data frequently used both as training data and as the source of documents to be landscaped.  The USPTO has recently leveraged such techniques in its own empirical studies on U.S. patent archives. 

Released in 2021, the USPTO's AI Patent Dataset identifies the presence of AI in over 13 million U.S. patent documents and further sorts them into one of eight component technologies~\citep{giczy2022}. This dataset was created by training a recurrent neural network in a semi-supervised manner to distinguish between positive and negative examples~\citep{abood2018}. The USPTO leveraged the AI Patent Dataset to trace the diffusion of AI and its component technologies within post-1976 U.S. patents~\citep{toole2020}, with such findings informing agency stakeholder engagements and other policy-relevant activities~\citep{uspto2022}.

Much patent analysis focuses on specifications, claims, and metadata, but an often-overlooked data source for patent analytics lies in prosecution history. The USPTO has applied AI techniques to make Office actions more accessible to the patent analysis community. Released in 2017, the USPTO's Office Action Research Dataset comprises a relational database of key elements from 4.4 million Office actions mailed during the 2008 to mid-2017 period~\citep{lu2017}. This dataset was created using machine learning and NLP techniques to systematically extract information from Office actions, thus marking the first time that comprehensive data on examiner-issued rejections was made readily available to the research community.
\section{Advancing the AI \& NLP research frontiers via USPTO data}

The foregoing body of work centers around the application of existing AI \& NLP techniques in IP-relevant areas. But another emerging body of work flips this paradigm by using USPTO data as an accelerant for scientific research in AI \& NLP. We highlight examples in both the training and evaluation of AI models.

\subsection{USPTO data in large language modeling}

Large language models have demonstrated a surprisingly diverse portfolio of natural language capabilities~\citep{devlin2019,brown2020}. Yet early iterations of billion-parameter language models employed lightly curated datasets constructed with few quality or diversity filters. Observing this, \citet{gao2020} compiled a dataset prioritizing both data quality and diversity, combining the background sections of millions of U.S. patents with 21 other data sources to form ``The Pile''.

This 825 GiB language modeling dataset, and subsets thereof, were subsequently used in training or assessing some of the largest and most advanced language models to appear in published research, including GPT-NeoX-20B, Gopher, and RETRO~\citep{black2022,rae2021,borgeaud2021}. OPT-175B~\citep{zhang2022}, currently the largest publicly-available language model, was trained on public USPTO data sourced from The Pile.

We observe that the background sections of patents, while informative, only scratch the surface of available content within the U.S. patent archives. Future language modeling datasets could include full patent specifications or prosecution history documents (\emph{e.g.}, Office actions). The latter holds particular promise as a source of scientific and legal reasoning examples not easily found elsewhere.

\subsection{Patent-sourced datasets for common tasks}

The quantity and detail of patent documents also readily enable their use as datasets for common AI \& NLP benchmark tasks. Patent classification (discussed in Section~\ref{sec:ai-nlp-ip-administration}) is, at its core, a quintessential multiclass classification challenge. AI researchers have already used public USPTO data and CPC annotations to create text classification benchmarks encompassing millions of patent documents~\citep{li2018, lee2020}. These benchmarks have been used to evaluate the capabilities of new self-attention neural network models~\citep{zaheer2020}.

Recent work has also augmented public USPTO data with automated data generation and manual annotations to form specialized benchmark datasets that can test the ability of novel AI and NLP models to penetrate complex technical concepts. For instance, \citet{aslanyan2022} construct a novel semantic similarity benchmark dataset by extrating phrases from patent documents, generating facially similar phrases, and manually rating the semantic similarity of each phrase pair on a five-point scale. A Kaggle competition featuring this benchmark resulted in nearly 43,000 submissions, achieving a top Pearson correlation of 87.8\%~\citep{kaggle2022}. The USPTO is interested in building upon these early successes by fostering future efforts to refashion patent data into valuable AI research benchmarks.
\section{Conclusion}

We have described two technical bodies of work that rest upon USPTO data. The first integrates USPTO data with AI \& NLP techniques to benefit IP administration, practice, and empirical analysis. The second leverages USPTO data in service of state-of-the-art AI \& NLP research.

We envision these two spheres forming a virtuous cycle wherein successes in one area furthers progress in the other. From search engines to benchmarks, and from landscapes to large language models and beyond, we hope that researchers and practitioners will find novel means of harnessing the richness of USPTO data to serve both the IP community and future AI researchers.
\begin{acks}
   We acknowledge our stellar colleagues from across the USPTO who have labored tirelessly to advance the agency's data dissemination and artificial intelligence efforts.
\end{acks}

\bibliographystyle{ACM-Reference-Format}
\bibliography{main}

\appendix

\end{document}